\documentclass[journal]{IEEEtran}

\usepackage{graphicx}
\usepackage{comment}
\usepackage{amsmath,amssymb} % define this before the line numbering.
\usepackage{color}

\usepackage{amsmath}
\usepackage{amssymb}

\usepackage{latexsym}
\usepackage{amsmath}
\usepackage{booktabs} % For formal tables
\usepackage{subfigure}
\usepackage{multirow}
\usepackage{algorithm}
\usepackage{listings}

\usepackage{etoolbox}
\makeatletter
\AfterEndEnvironment{algorithm}{\let\@algcomment\relax}
\AtEndEnvironment{algorithm}{\kern2pt\hrule\relax\vskip3pt\@algcomment}
\let\@algcomment\relax
\newcommand\algcomment[1]{\def\@algcomment{\footnotesize#1}}
\renewcommand\fs@ruled{\def\@fs@cfont{\bfseries}\let\@fs@capt\floatc@ruled
  \def\@fs@pre{\hrule height.8pt depth0pt \kern2pt}%
  \def\@fs@post{}%
  \def\@fs@mid{\kern2pt\hrule\kern2pt}%
  \let\@fs@iftopcapt\iftrue}
\makeatother
\begin{document}
\title{Cross-Modal Food Retrieval: Learning a Joint Embedding of Food Images and Recipes with Semantic Consistency and Attention Mechanism}

\author{Hao~Wang\thanks{Work done in Singapore Management University.},
        Doyen~Sahoo,
        Chenghao~Liu,
        Ke~Shu,
        Palakorn~Achananuparp,
        Ee-peng~Lim,
        and~Steven~C.~H.~Hoi,~\IEEEmembership{Fellow,~IEEE}% <-this % stops a space
\thanks{Hao Wang is with Nanyang Technological University; e-mail: hao005@ntu.edu.sg.}% <-this % stops a space
\thanks{Doyen Sahoo and Chenghao Liu are with Salesforce Research Asia; e-mail: \{doyensahoo, twinsken\}@gmail.com} % <-this % stops a space
\thanks{Ke Shu, Palakorn Achananuparp, and Ee-peng Lim are with Singapore Management University; e-mail: \{keshu, palakorna, eplim\}@smu.edu.sg.}
\thanks{Steven C. H. Hoi is with Singapore Management University and Salesforce Research Asia; e-mail: chhoi@smu.edu.sg.}% <-this % stops a space
}


\markboth{}%
% \markboth{Journal of \LaTeX\ Class Files,~Vol.~14, No.~8, August~2015}%
{Shell \MakeLowercase{\textit{et al.}}: Bare Demo of IEEEtran.cls for IEEE Journals}



\maketitle

\begin{abstract}
   Food retrieval is an important task to perform analysis of food-related information, where we are interested in retrieving relevant information about the queried food item such as ingredients, cooking instructions, etc. 
   In this paper, we investigate cross-modal retrieval between food images and cooking recipes. The goal is to learn an embedding of images and recipes in a common feature space, such that the corresponding image-recipe embeddings lie close to one another.
   Two major challenges in addressing this problem are 1) large intra-variance and small inter-variance across cross-modal food data; and 2) difficulties in obtaining discriminative recipe representations. To address these two problems, we propose Semantic-Consistent and Attention-based Networks (SCAN), which regularize the embeddings of the two modalities through aligning output semantic probabilities. Besides, we exploit a self-attention mechanism to improve the embedding of recipes. We evaluate the performance of the proposed method on the large-scale Recipe1M dataset, and show that we can outperform several state-of-the-art cross-modal retrieval strategies for food images and cooking recipes by a significant margin.
\end{abstract}

\begin{IEEEkeywords}
Deep Learning, Cross-Modal Retrieval, Vision-and-Language.
\end{IEEEkeywords}

% \fi
\IEEEpeerreviewmaketitle

\begin{figure*}[htb]
\begin{center}
\includegraphics[width=0.95\textwidth]{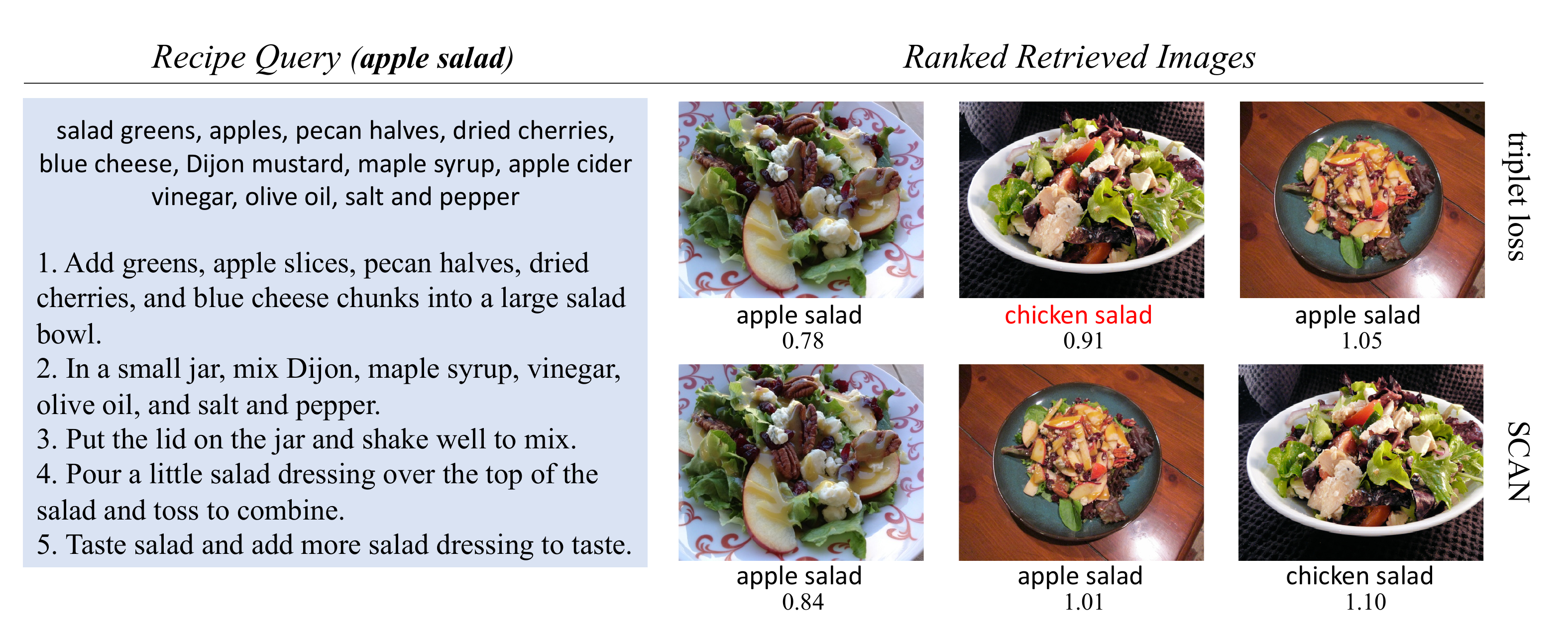}
\end{center}
   \caption{\textbf{Recipe-to-image retrieval ranked results:} Take an \emph{apple salad} recipe as the query, which contains ingredients and cooking instructions, we show the retrieval results based on \textbf{Euclidean distance} (as the numbers indicated in the figure) for 3 different food images with large intra-class variance and small inter-class variance, i.e. images of \emph{apple salad} have different looks, while \emph{chicken salad} image is more similar to \emph{apple salad}. We rank the retrieved results of using (i) vanilla triplet loss; and (ii) our proposed SCAN model. It shows vanilla triplet loss outputs a wrong ranking order, while SCAN can provide more precise ranking results.}
\label{fig:motivation}
\end{figure*}

\section{Introduction}
%In recent years, cross-modal retrieval has drawn much attention with the rapid growth of multimodal data. 
Food plays an essential role in human daily life. To discover the relationship between cross-modal food data, i.e. food images and recipes, we aim to address the problem of cross-modal food retrieval based on a large amount of heterogeneous food dataset \cite{salvador2017learning}. Specifically, we take the cooking recipes (ingredients \& cooking instructions) as the query to retrieve the food images, and vice versa.
Recently, there have been many works \cite{salvador2017learning,carvalho2018cross,wang2019learning} on cross-modal food retrieval, where they mainly learn the joint embeddings for recipes and images with vanilla pair-wise loss to achieve the cross-modal alignment. Despite those efforts, cross-modal food retrieval remains challenging mainly due to the following two reasons: 1) the large intra-class variance across food data pairs, and 2) the difficulties of obtaining discriminative recipe representation. 

In cross-modal food data, given a recipe, we may have many food images that are cooked by different chefs. Besides, the images from different recipes can look very similar because they have similar ingredients. Hence, the data representation from the same food can be different, but different food may have similar data representations. This leads to large intra-class variance but small inter-class variance in food data. Existing studies \cite{carvalho2018cross,chen2018deep,chen2017cross} only address the small inter-class variance problem by utilizing triplet loss to measure the similarities between cross-modal data. Specifically, the objective of triplet loss is to make inter-class feature distance larger than intra-class feature distance by a predefined margin \cite{cheng2016person}. Therefore, cross-modal instances from the same class may form a loose cluster with a large average intra-class distance. As a consequence, it eventually results in less-than-optimal ranking, i.e., irrelevant images are closer to the queried recipe than relevant images. See Figure \ref{fig:motivation} for an example.

Besides, many recipes share common ingredients in different food. For instance \emph{fruit salad} has ingredients of \emph{apple}, \emph{orange} and \emph{sugar} etc., where \emph{apple} and \emph{orange} are the main ingredients, while \emph{sugar} is one of the ingredients in many other foods. If the embeddings of ingredients are treated equally during training, the features learned by the model may not be discriminative enough. In addition, the cooking instructions crawled from cooking websites tend to be noisy, some instructions turn out irrelevant to cooking e.g. \emph{'Enjoy!'}, which convey no information for cooking instruction features but degrade the performance of cross-modal retrieval task. In order to find the attended ingredients, Chen et al. \cite{chen2017cross} apply a two-layer deep attention mechanism, which learns joint features by locating the visual food regions that correspond to ingredients. However, this method relies on high-quality food images and essentially increases the computational complexity.

To resolve those issues, we propose a novel unified framework of \textbf{S}emantic-\textbf{C}onsistent and \textbf{A}ttention-Based \textbf{N}etwork (SCAN) to improve the cross-modal food retrieval performance. The pipeline of the framework is shown in Figure \ref{fig:pipeline}. To reduce the intra-class variance, we introduce a semantic consistency loss, which imposes Kullback-Leibler (KL) Divergence to minimize the difference between the output semantic probabilities of paired image and recipe, such that the image and recipe representations would follow similar distributions. In order to obtain discriminative recipe representations, we combine the self-attention mechanism \cite{vaswani2017attention} with LSTM to find the key ingredients and cooking instructions for each recipe. Without requiring food images or adding extra layers, we can learn better discriminative recipe embeddings, compared to that trained with plain LSTM. 

Our work makes two major contributions as follows:
\begin{itemize}
   \item We introduce a semantic consistency loss to cross-modal food retrieval task. The result shows that it can align cross-modal matching pairs and reduce the intra-class variance of food data representations.
   \item We integrate the self-attention mechanism with LSTM, and learn discriminative recipe features without requiring the food images. It is useful to discriminate samples of similar recipes.
\end{itemize}

We perform extensive experimental analysis on Recipe1M, which is the largest cross-modal food dataset and available in the public. We find that our proposed cross-modal food retrieval approach SCAN outperforms state-of-the-art methods. Finally, we show some visualizations of the retrieved results.

\section{Related Work}
\subsection{Cross-modal Retrieval}
Our work is closely related to the general cross-modal retrieval task, which aims to retrieve the corresponding instance of different modalities based on the given query. The general idea of cross-modal retrieval is to correlate heterogeneous data, mapping the data from different modalities to the common space. As an early work for multi-media, Canonical Correlation Analysis (CCA) \cite{hotelling1936relations} utilizes global alignment to allow the data mapping of different modalities with similar semantics to be close in the common space, by maximizing the correlation between cross-modal similar pairs. However, CCA-based approaches model the cross-modal data only by linear projections, when it comes to large-scale complex real-world data, it is difficult for CCA to fully model the correlations.

Many recent works \cite{li2018deep,li2020weakly,jin2020deep} utilize deep architectures for cross-modal retrieval, which have the advantage of capturing complex non-linear cross-modal correlations. Specifically, to improve the efficiency in retrieval process, hashing has been introduced to multimedia retrieval \cite{li2020weakly,jin2020deep}. With the advent of generative adversarial networks (GANs) \cite{goodfellow2014generative}, which are helpful to model the data distributions, some adversarial training methods \cite{gu2017look, peng2017cm, ghasedi2018unsupervised} are frequently used for modality fusion. Peng et al. \cite{peng2017cm} utilize two kinds of discriminative models to simultaneously conduct intra-modality and inter-modality discrimination, and model the joint distribution over the data of different modalities. \cite{gu2017look,jing2020incomplete} incorporate generative processes into the cross-modal feature embedding. Specifically, Gu et al. \cite{gu2017look} try to generate the images from text features and generate corresponding captions from the image features. In this way, they learn not only the global abstract features but also the local grounded features. To address the challenge that unpaired data may exist in the cross-modal dataset, Jing et al. \cite{jing2020incomplete} propose to learn modality-invariant representations with autoencoders, which are further dual-aligned at the distribution level and the semantic level. To learn fine-grained phrase correspondence, Liu et al. \cite{liu2020graph} construct textual and visual graph, they learn the cross-modal correspondence by node-level and structure-level matching.

\begin{figure*}[htb]
\begin{center}
\includegraphics[scale=0.47]{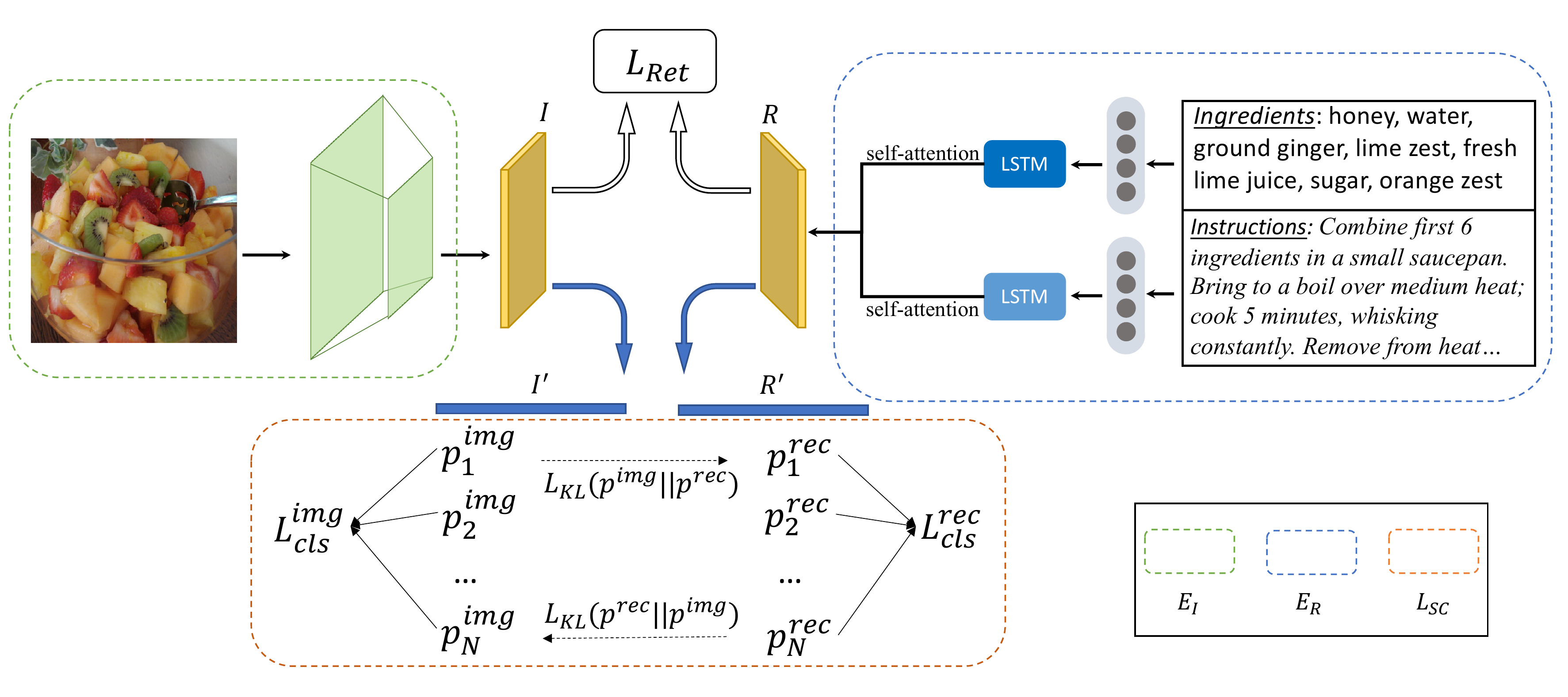}
\end{center}
   \caption{Our proposed framework for cross-modal retrieval task. We have two branches to encode food images and recipes respectively. One embedding function $E_I$ is designed to extract food image representations $I$, where a CNN is used. The other embedding function $E_R$ is composed of two LSTMs with self-attention mechanism, designed for obtaining discriminative recipe representations $R$. $I$ and $R$ are fed into retrieval loss (triplet loss) $L_{Ret}$ to do cross-modal retrieval learning. We add another FC transformation on $I$ and $R$ with the output dimensionality as the number of food categories, to obtain the semantic probabilities $p^{img}$ and $p^{rec}$, where we utilize semantic consistency loss $L_{SC}$ to correlate food image and recipe data.}
\label{fig:pipeline}
\end{figure*}

\subsection{Food Computing}
Food computing \cite{min2018survey} utilizes computational methods to analyze the food data including the food images and recipes. Many food-related computational tasks have been widely researched, like food recognition \cite{min2017you, horiguchi2018personalized,jiang2019multi,min2020isia}, retrieval \cite{min2016being, salvador2017learning}, recommendation \cite{ge2015using,min2019food} and recipe generation \cite{wang2020structure}, etc. In this paper, we mainly investigate the cross-modal food retrieval problem based on Recipe1M dataset \cite{salvador2017learning}.

Recipe1M \cite{salvador2017learning} is currently the largest cross-modal food dataset, which was scraped from over two dozen popular cooking websites, and contains rich cooking instructions, ingredient information, and the corresponding cooked food images. Besides, half amount of the data in Recipe1M has semantic food category labels, which are extracted from the food titles in the cooking websites. Recipe1M is proposed mainly for the cross-modal food retrieval task.

\cite{chen2016deep, min2016being} are early works in cross-modal food retrieval. In \cite{chen2016deep}, a multi-task deep learning architecture is proposed for simultaneous ingredient and food recognition. The learned visual features and semantic attributes of ingredients are then used for recipe retrieval, but they only test their model in a small-scale dataset, and cannot demonstrate the efficacy in real-world large-scale data. Min et al. \cite{min2016being} utilize a multi-modal Deep Boltzmann Machine for recipe-image retrieval. \cite{chen2017cross, chen2018deep} integrate attention mechanism into cross-modal retrieval, Chen et al. \cite{chen2017cross} introduce a stacked attention network (SAN) to learn joint space from images and recipes for cross-modal retrieval. However, SAN only considers ingredient lists and ignores the rich information provided by cooking instructions, so they have poor performance in Recipe1M dataset. Consequently, Chen et al. improve the previous work SAN in \cite{chen2018deep}, where they make full use of the ingredient, cooking instruction, and title (food category) information of Recipe1M, and concatenate the three types of features above to construct the recipe embeddings. Compared with the self-attention mechanism we adopt in our model, both \cite{chen2017cross} and \cite{chen2018deep} add extra learnable parameters to compute the attended parts,
% adopt a two-layer deep attention mechanism to learn the embedding space, 
% and \cite{chen2018deep} adds an extra learnable context vector to compute the attended parts, 
which increase the computational complexity. In order to have better regularization on the shared representation space learning, \cite{salvador2017learning, carvalho2018cross} both corporate the semantic labels with the joint training. Salvador et al. \cite{salvador2017learning} develop a hybrid neural network architecture with a cosine embedding loss for retrieval learning and a cross-entropy loss for classification, such that a joint common space for image and recipe embeddings can be learned for cross-modal retrieval. \cite{carvalho2018cross} is an extended version of \cite{salvador2017learning}, providing a double-triplet strategy to express both the retrieval loss and the classification loss.

Different from existing cross-modal food retrieval work, we propose a novel semantic consistency loss with a self-attention mechanism, where we impose regularization on the output semantic probabilities of paired food image and recipe embeddings, to correlate the learned food image and recipe representations. Self-attention helps learn discriminative recipe embeddings without depending on the food images or adding some extra learnable parameters.

\section{Proposed Methods}
In this section, we introduce our proposed model, where we utilize food image-recipe paired data to learn cross-modal embeddings as shown in Figure \ref{fig:pipeline}.

\subsection{Overview}
We formulate the proposed cross-modal food retrieval with three networks, i.e. one convolutional neural network (CNN) for food image embeddings, and two LSTMs to encode ingredients and cooking instructions respectively. The food image representations $I$ can be obtained from the output of CNN directly, while the recipe representations $R$ come from the concatenation of the ingredient features $f_{ingredient}$ and instruction features $f_{instruction}$. 
Specifically, for obtaining discriminative ingredient and instruction embeddings, we integrate the self-attention mechanism \cite{vaswani2017attention} into the LSTM embedding. 
Triplet loss is used as the main loss function $L_{Ret}$ to map cross-modal data to the common space, and semantic consistency loss $L_{SC}$ is utilized to align cross-modal matching pairs for retrieval task, reducing the intra-class variance of food data. The overall objective function of the proposed SCAN is given as:
\begin{equation}
\begin{aligned}
L = L_{Ret} + \lambda L_{SC},
\label{total_eq}
\end{aligned}
\end{equation}

\begin{figure}
\begin{center}
\includegraphics[width=0.5\textwidth]{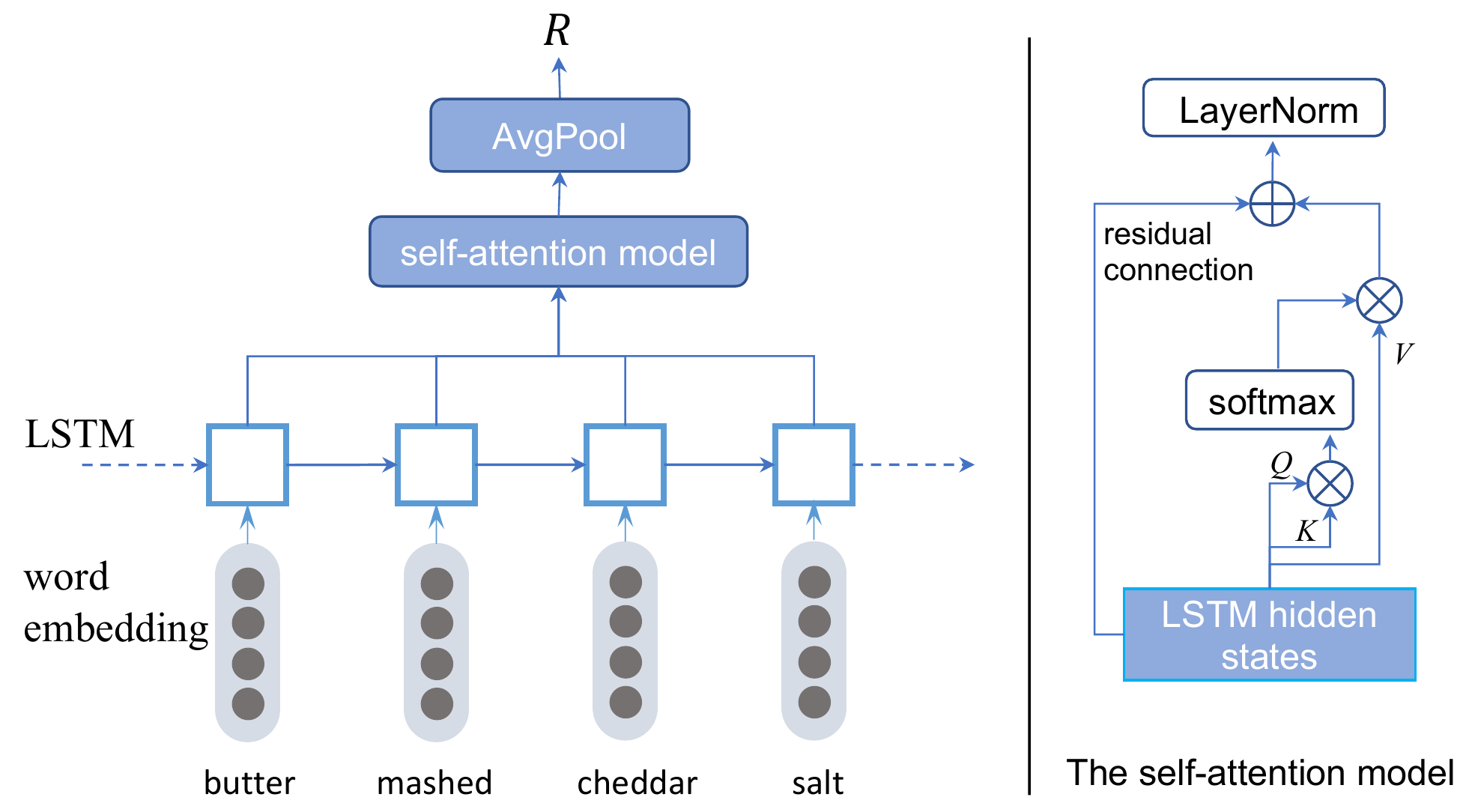}
\end{center}
   \caption{The structure of ingredient (instruction) embedding model with self-attention mechanism. $Q$, $K$ and $V$ denote queries, keys and values respectively.}
\label{fig:attn_model}
\end{figure}

\subsection{Recipe Embedding}
We use two LSTMs to get ingredient and instruction representations $f_{ingredient}$, $f_{instruction}$, concatenate them and pass through a fully-connected layer to give a 1024-dimensional feature vector, as the recipe representation $R$.

\subsubsection{Ingredient Representation Learning}
Instead of word-level word2vec representations, ingredient-level word2vec representations are used in ingredient embedding. To be specific, \emph{ground ginger} is regarded as a single word vector, instead of two separate word vectors of \emph{ground} and \emph{ginger}. 

We integrate the self-attention mechanism with LSTM output to construct recipe embeddings. The purpose of applying the self-attention model lies in assigning higher weights to main ingredients for different food items, making the attended ingredients contribute more to the ingredient embedding, while reducing the effect of common ingredients. The self-attention structure is shown in Figure \ref{fig:attn_model}.

Given an ingredient input $\{z_1, z_2, ..., z_n\}$, we first encode it with pretrained embeddings from word2vec algorithm to obtain the ingredient representation $Z_t$. Then $\{Z_1, Z_2, ..., Z_n\}$ will be fed into the one-layer bidirectional LSTM as a sequence step by step. For each step $t$, the recurrent network takes in the ingredient vector $Z_t$ and the output of previous step $h_{t-1}$ as the input, and produces the current step output $h_t$ by a non-linear transformation, as follow:

\begin{equation}
\begin{aligned}
h_t = \mathrm{tanh}(\textbf{W}{Z_t} + \textbf{U}{h_{t-1}} + b),
\end{aligned}
\end{equation}

The bidirectional LSTM consists of a forward hidden state $\overrightarrow{h_t}$ which processes ingredients from $Z_1$ to $Z_n$ and a backward hidden state $\overleftarrow{h_t}$ which processes ingredients from $Z_n$ to $Z_1$. We obtain the representation $h_t$ of each ingredient $z_t$ by concatenating $\overrightarrow{h_t}$ and $\overleftarrow{h_t}$, i.e. $h_t=[\overrightarrow{h_t}, \overleftarrow{h_t}]$, so that the representation of the ingredient list of each food item is $H = \{h_1, h_2, ..., h_n\}$.

We further measure the importance of ingredients in the recipe with the self-attention mechanism which has been studied in Transformer \cite{vaswani2017attention}, where the input comes from queries $Q$ and keys $K$ of dimension $d_k$, and values $V$ of dimension $d_v$ (the definition of $Q$, $K$ and $V$ can be referred in \cite{vaswani2017attention}), we compute the attention output as:

\begin{equation}
\begin{aligned}
\mathrm{Attention}(Q, K, V) = \mathrm{softmax}(\frac{QK^T}{\sqrt{d_k}})V,
\end{aligned}
\end{equation}

Different from the earlier attention-based methods \cite{chen2018deep}, we use self-attention mechanism where all of the keys, values and queries come from the same ingredient representation $H$. Therefore, the computational complexity is reduced since it is not necessary to add extra layers to train attention weights. The ingredient attention output $H_{attn}$ can be formulated as:

\begin{equation}
\begin{aligned}
H_{attn} = \mathrm{Attention}(H, H, H) \\
= \mathrm{softmax}(\frac{HH^T}{\sqrt{d_h}})H,
\end{aligned}
\end{equation}
where $d_h$ is the dimension of $H$. In order to enable unimpeded information flow for recipe embedding, skip connections are used in the attention model. Layer normalization \cite{ba2016layer} is also used since it is effective in stabilizing the hidden state dynamics in recurrent network. The final ingredient representation $f_{ingredient}$ is generated from summation of $H$ and $H_{attn}$, which can be defined as:

\begin{equation}
\begin{aligned}
f_{ingredient} = \mathrm{LayerNorm}(H_{attn} + H),
\end{aligned}
\end{equation}

\subsubsection{Instruction Representation Learning}
Considering that cooking instructions are composed of a sequence of variable-form and lengthy sentences, we compute the instruction embedding with a two-stage LSTM model. For the first stage, we apply the same approach as \cite{salvador2017learning} to obtain the representations of each instruction sentence, in which it uses skip-instructions \cite{salvador2017learning} with the technique of skip-thoughts \cite{kiros2015skip}. 

The next stage is similar to the ingredient representation learning. We feed the pre-computed fixed-length instruction sentence representation into the LSTM model to generate the hidden representation of each cooking instruction sentence. Based on that, we can obtain the self-attention representation. The final instruction feature $f_{instruction}$ is generated from the layer normalization function on the previous two representations, as we formulate in the last section. By doing so, we are able to find the key sentences in cooking instruction. Some visualizations on attended ingredients and cooking instructions can be found in Section \ref{attention_vis}.

\subsection{Image Embedding}
We use ResNet-50 \cite{he2016deep} pretrained on ImageNet to encode food images. The dimension of the final food image features $I$ is 1024, which is identical to that of recipe features $R$.

\subsection{Cross-modal Food Retrieval Learning}
Triplet loss is utilized to do retrieval learning, the objective function is:

\begin{equation}
\begin{aligned}
L_{Ret} =  & \sum_I\left[d(I_{a}, R_{p})-d(I_{a}, R_{n})+\alpha\right]_+ \; \\
& + \sum_R\left[d(R_{a}, I_{p})-d(R_{a}, I_{n})+\alpha\right]_+,
\end{aligned}
\end{equation}
where $d(\bullet)$ is the Euclidean distance, subscripts $a,p$ and $n$ refer to anchor, positive and negative samples respectively and $\alpha$ is the margin. The summation symbol means that we construct triplets and do the training for all samples in the mini-batch. To improve the effectiveness of training, we adopt the \emph{BatchHard} idea proposed in \cite{hermans2017defense}. Specifically in a mini-batch, each sample can be used as an anchor, then for each anchor, we select the closest negative sample and the farthest positive sample to construct the triplet.
% given an anchor sample, we select the most distant positive instance and the closest negative instance, to construct the triplet.

\subsection{Semantic Consistency}

Given the pairs of food image and recipe representations $I$, $R$, we first transform $I$ and $R$ into $I'$ and $R'$ for classification with an extra FC layer. The dimension of the output is same as the number of categories $N$. The probabilities of food category $i$ can be computed by a softmax activation as:
 
\begin{equation}
\begin{aligned}
p_{i}^{img} = \frac{exp(I_{i}')}{\sum_{i=1}^{N} exp(I_{i}')}, 
\end{aligned}
\end{equation}

\begin{equation}
\begin{aligned}
p_{i}^{rec} = \frac{exp(R_{i}')}{\sum_{i=1}^{N} exp(R_{i}')}, 
\end{aligned}
\end{equation}
where $N$ represents the total number of food categories. Given $p_{i}^{img}$ and $p_{i}^{rec}$, where $i \in \{1, 2, ..., N\}$, the predicted label $l^{img}$ and $l^{rec}$ for each food item can be obtained. We formulate the classification (cross-entropy) loss as $L_{cls}(p^{img}, p^{rec}, c^{img}, c^{rec})$, where $c^{img}$, $c^{rec}$ are the ground-truth class label for food image and recipe respectively.

In the prior work \cite{salvador2017learning}, $L_{cls}(p^{img}, p^{rec}, c^{img}, c^{rec})$ consists of $L^{img}_{cls}$ and $L^{rec}_{cls}$, which are treated as two independent classifiers, focusing on the regularization on the embeddings from food images and recipes separately. However, food image and recipe embeddings come from heterogeneous modalities, the output probabilities of each category can be significantly different, i.e. for each food item, the distributions of $p_{i}^{img}$ and $p_{i}^{rec}$ remain big variance. As a result, the distance of intra-class features remains large. To improve image-recipe matching and make the probabilities predicted by different classifiers consistent, we minimize Kullback-Leibler (KL) Divergence between $p_{i}^{img}$ and $p_{i}^{rec}$ of paired cross-modal data for each food item, which can be formulated as:

\begin{equation}
\begin{aligned}
 L_{KL}({p}^{img}\|{p}^{rec}) = \sum_{i=1}^{N} \ p_i^{img} \log \frac{p_i^{img}}{p_i^{rec}},
\end{aligned}
\end{equation}

\begin{equation}
\begin{aligned}
 L_{KL}({p}^{rec}\|{p}^{img}) = \sum_{i=1}^{N} \ p_i^{rec} \log \frac{p_i^{rec}}{p_i^{img}},
\end{aligned}
\end{equation}

By aligning the output probabilities of cross-modal data representations, we minimize the intra-class variance with back-propagation. The overall semantic consistency loss $L_{SC}$ is defined as:

\begin{equation}
\begin{aligned}
 L_{SC} = \{( L^{img}_{cls} + L_{KL}({p}^{rec}\|{p}^{img})) \; \\
 + ( L^{rec}_{cls}+ L_{KL}({p}^{img}\|{p}^{rec}))\} / 2.
\end{aligned}
\end{equation}

%##################################################################################################
\begin{algorithm}[t]
\caption{Pseudocode of SCAN in a PyTorch-like style.}
\label{alg:code}
\algcomment{\fontsize{7.2pt}{0em}\selectfont \texttt{sqrt}: square root; \texttt{bmm}: batch matrix multiplication; \texttt{cat}: concatenation.
%\vspace{-1.em}
}
\definecolor{codeblue}{rgb}{0.25,0.5,0.5}
\lstset{
  backgroundcolor=\color{white},
  basicstyle=\fontsize{7.2pt}{7.2pt}\ttfamily\selectfont,
  columns=fullflexible,
  breaklines=true,
  captionpos=b,
  commentstyle=\fontsize{7.2pt}{7.2pt}\color{codeblue},
  keywordstyle=\fontsize{7.2pt}{7.2pt},
%  frame=tb,
}
\begin{lstlisting}[language=python]
# load a minibatch containing ingredients, instructions and images with N samples
for (ingr, instr, img) in loader:
    
    # perform word embedding on ingredients
    Z = Embedding.forward(ingr)
    # compute LSTM features, Eq. (2)
    H = LSTM.forward(Z)
    # set the temperate
    t = sqrt(dimension_H)
    # compute attention scores 
    attn = bmm(H, H.T) / t
    # compute attention outputs, Eq. (4)
    output = bmm(attn, H)
    # use residual connection to get the final self-attention outputs, Eq. (5)
    f_ingredient = LayerNorm(output + H)
    
    # use self-attention to get the instruction features
    f_instruction = SelfAttention.forward(instr)
    
    # compute the recipe features
    R = cat([f_ingredient, f_instruction], dim=1)
    # compute the image features
    I = CNN.forward(img)
    # compute triplet loss, Eq. (6)
    L_Ret = TripletLoss(R, I)
    
    # compute the class probabilities for R and I, Eq. (7, 8)
    p_rec = softmax(R)
    p_img = softmax(I)
    # compute cross-entropy loss
    L_cls_rec = CrossEntropyLoss(p_rec, labels)
    L_cls_img = CrossEntropyLoss(p_img, labels)
    L_cls = (L_cls_rec + L_cls_img) / 2
    # compute KL divergence between recipes and images, Eq. (9, 10)
    L_KL = (KL(p_rec||p_img) + KL(p_img||p_rec)) / 2 
    # compute the semantic consistency loss, Eq. (11)
    L_SC = L_cls + L_KL
    
    # Eq. (1)
    loss = L_Ret + L_SC

    # Adam update
    loss.backward()
\end{lstlisting}
\end{algorithm}
%##################################################################################################

\begin{table*}[h!]
  \centering
    \caption{\textbf{Main Results.} Evaluation of the performance of our proposed method compared against the baselines. The models are evaluated on the basis of MedR, where lower is better, and R@K (\%), where higher is better.}
  \begin{tabular}{clcccccccc}
  \toprule
   \textbf{Size of Test Set} && \multicolumn{4}{c}{\textbf{Image-to-Recipe Retrieval}} & \multicolumn{4}{c}{\textbf{Recipe-to-Image Retrieval} }\\
    \midrule
    \multicolumn{1}{c}{} & \textbf{Methods} & \textbf{medR $\downarrow$} & \textbf{R@1 $\uparrow$} &\textbf{R@5 $\uparrow$}& \textbf{R@10 $\uparrow$} &  \textbf{medR $\downarrow$} & \textbf{R@1 $\uparrow$} & \textbf{R@5 $\uparrow$}& \textbf{R@10 $\uparrow$} \\
    \midrule
	\multirow{10}{*}{{1k}} 
        &CCA \cite{hotelling1936relations}& 15.7 & 14.0 & 32.0 & 43.0 & 24.8 & 9.0 & 24.0 & 35.0 \\
        &SAN \cite{chen2017cross}& 16.1 & 12.5 & 31.1 & 42.3 & - & - & - & -\\
        &JE \cite{salvador2017learning}& 5.2 & 24.0 & 51.0 & 65.0 & 5.1 & 25.0 & 52.0 & 65.0  \\
        &AM \cite{chen2018deep}& 4.6 & 25.6 & 53.7 & 66.9 & 4.6 & 25.7 & 53.9 & 67.1  \\
        &AdaMine \cite{carvalho2018cross}& 2.0 & 39.8 & 69.0 & 77.4 & 1.0 & 40.2 & 68.1 & 78.7 \\
        &$\rm {R^2 GAN}$ ~\cite{zhu2019r2gan} & 2.0 & 39.1 & 71.0 & 81.7 & 2.0 & 40.6 & 72.6 & 83.3 \\
        &MCEN \cite{fu2020mcen} & 2.0 & 48.2 & 75.8 & 83.6 & 2.0 & 48.4 & 76.1 & 83.7 \\
        &ACME \cite{wang2019learning} & 1.0 & 51.8 & 80.2 & 87.5 & 1.0 & 52.8 & 80.2 & 87.6 \\
        &tri-pro \cite{zan2020sentence} & 1.0 & 52.7 & 81.0 & 88.1 & 1.0 & 53.8 & 81.1 & 88.3 \\
        &SCAN (Ours) & 1.0 & \textbf{54.0}  & \textbf{81.9}  & \textbf{89.2}  & 1.0 & \textbf{54.9}  & \textbf{81.9}  & \textbf{89.0} \\
    \midrule
	\multirow{8}{*}{{10k}} 
	& JE \cite{salvador2017learning} & 41.9 & - & - & - & 39.2 & - & - & - \\
	&AM \cite{chen2018deep}& 39.8 & 7.2 & 19.2 & 27.6 & 38.1 & 7.0 & 19.4 & 27.8  \\ 
    &AdaMine \cite{carvalho2018cross}& 13.2 & 14.9 & 35.3 & 45.2 & 12.2 & 14.8 & 34.6 & 46.1\\
    &$\rm {R^2 GAN}$ \cite{zhu2019r2gan} & 13.9 & 13.5 & 33.5 & 44.9 & 11.6 & 14.2 & 35.0 & 46.8 \\
    &MCEN \cite{fu2020mcen} & 7.2 & 20.3 & 43.3 & 54.4 & 6.6 & 21.4 & 44.3 & 55.2 \\
    &tri-pro \cite{zan2020sentence} & 7.0 & 22.1 & 45.9 & 56.9 & 7.0 & 23.4 & 47.3 & 57.9 \\
    &ACME \cite{wang2019learning} & 6.7 & 22.9 & 46.8 & 57.9 & 6.0 & 24.4 & 47.9 & 59.0 \\
    & SCAN (Ours) & 5.9 & \textbf{23.7} & \textbf{49.3} & \textbf{60.6} & 5.1 & \textbf{25.3} & \textbf{50.6} & \textbf{61.6}\\
    \bottomrule
  \end{tabular}
  \label{tab:results}
\end{table*}

\begin{figure*}[h!]
\begin{center}
	\subfigure{
		\includegraphics[width=0.32\textwidth]{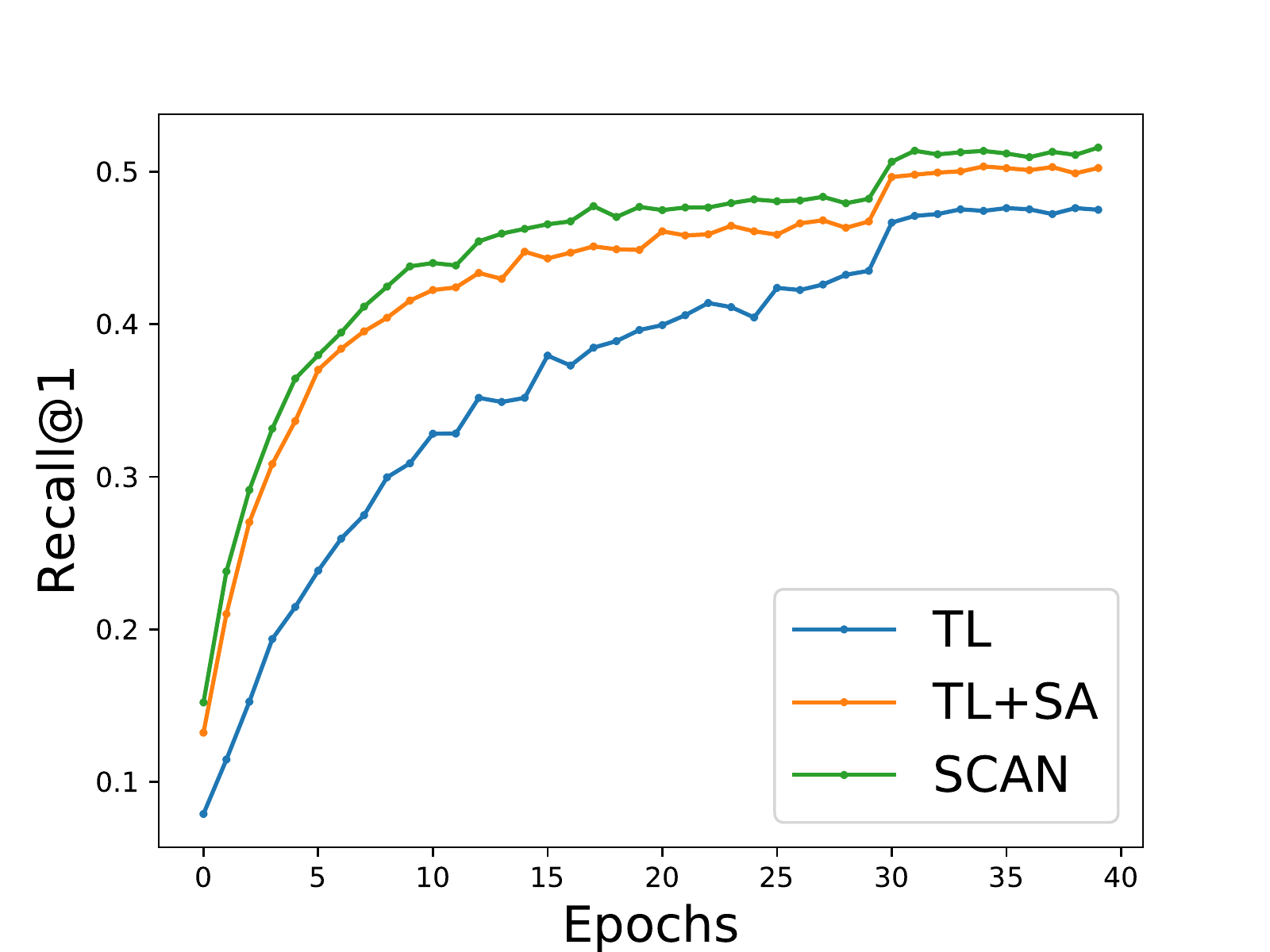}}
	\subfigure{
		\includegraphics[width=.32\textwidth]{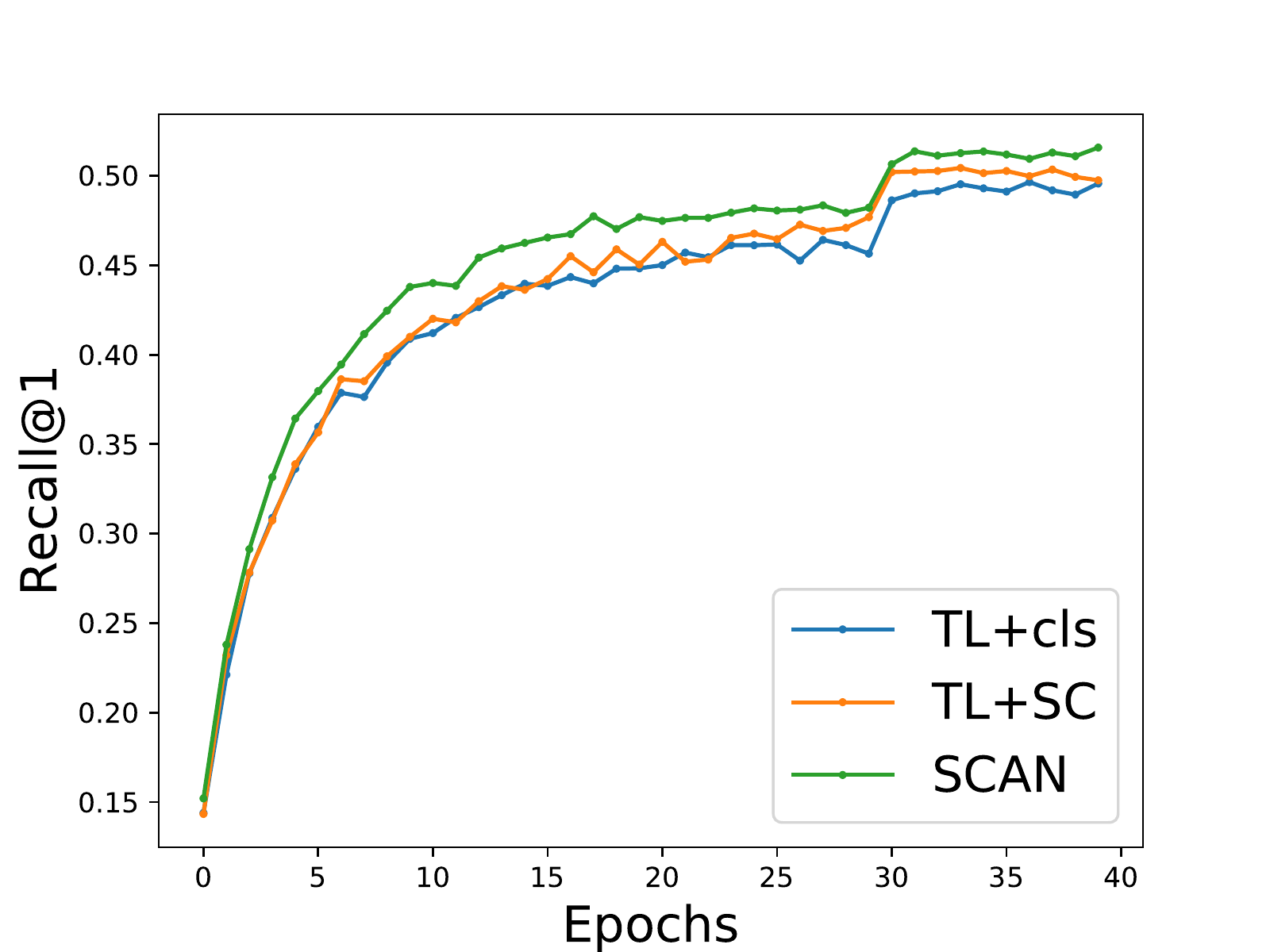}}
	\subfigure{
		\includegraphics[width=.32\textwidth]{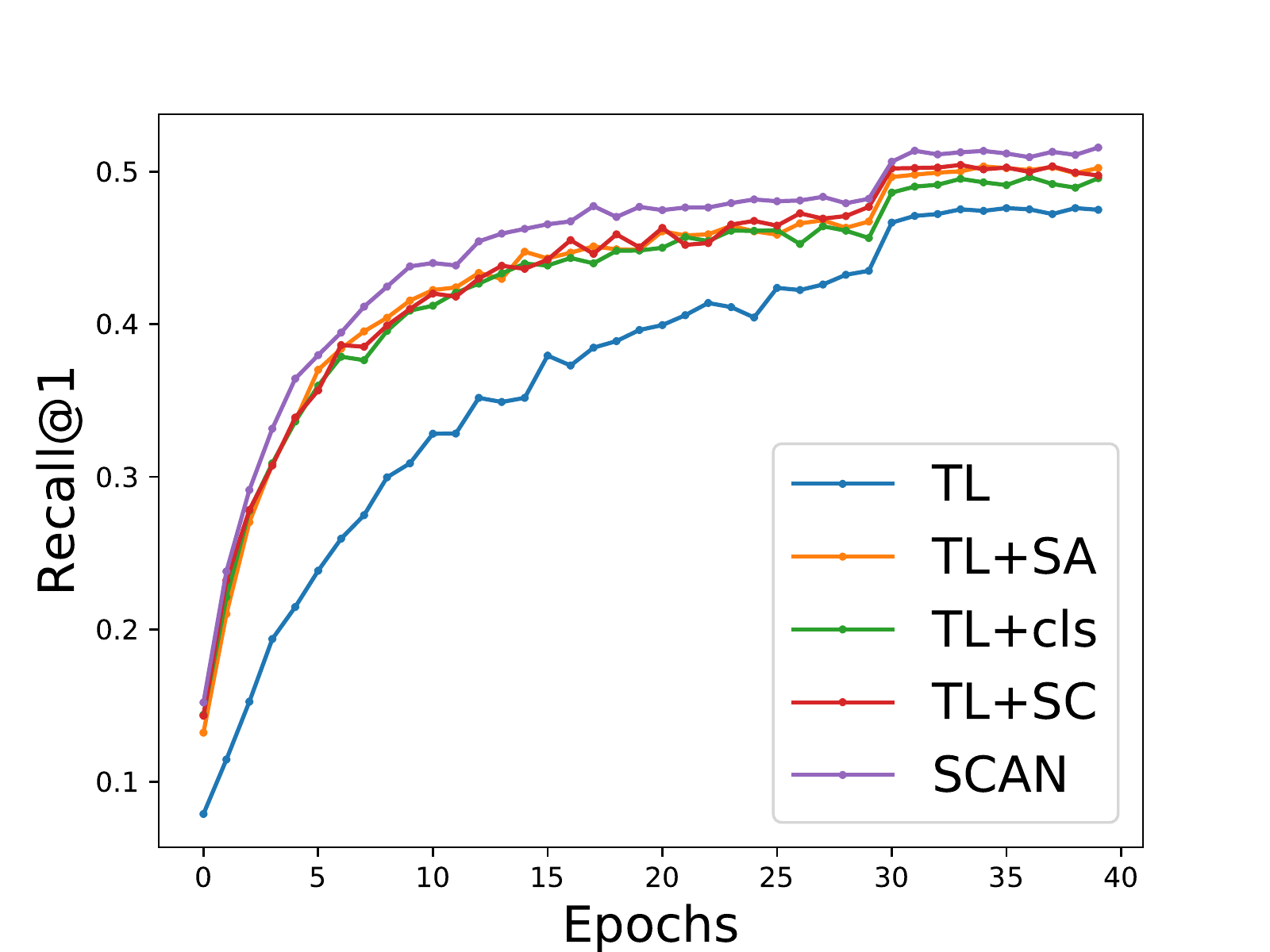}}
\end{center}
\caption{The training records of our proposed model SCAN and each component of SCAN.}
\label{fig:training}
\end{figure*}

\begin{table}
  \centering
  \caption{The performance of proposed model SCAN trained with different trade-off parameter $\lambda$.}
    \begin{tabular}{lcccc}
    \toprule
    {$\lambda$} & \textbf{MedR} & \textbf{R@1 (\%)} & \textbf{R@5 (\%)} & \textbf{R@10 (\%)} \\
    \midrule
    \textbf{0.01} & 1.0 & 53.2 & 81.3  & 88.2 \\
    \textbf{0.05} & 1.0 & \textbf{54.0}  & \textbf{81.9}  & \textbf{89.2} \\
    \textbf{0.1} & 1.0 & 51.9  & 80.5  & 87.6 \\
    \textbf{0.5} & 1.7 & 42.7  & 71.8  & 81.1 \\
    \bottomrule
    \end{tabular}%
  \label{tab:parameter}%
\end{table}%

\section{Experiments}

\subsection{Dataset}\label{Dataset}
We conduct extensive experiments to evaluate the performance of our proposed methods in Recipe1M dataset \cite{salvador2017learning}, the largest cooking dataset with recipe and food image pairs available to the public. Recipe1M was scraped from over 24 popular cooking websites and it not only contains the image-recipe paired labels but also more than half the amount of the food data with semantic category labels extracted from food titles on the websites. The category labels provide semantic information for the cross-modal retrieval task, making it fit in our proposed method well. The paired labels and category labels construct the hierarchical relationships among the food. One food category (e.g. \emph{fruit salads}) may contain hundreds of different food pairs, since there are many recipes of different \emph{fruit salads}.

We perform the cross-modal food retrieval task based on food data pairs, i.e. when we take the recipes as the query to do the retrieval, the ground truth will be the food images in food data pairs, and vice versa. We use the original Recipe1M data split \cite{salvador2017learning}, containing 238,999 image-recipe pairs for training, 51,119 and 51,303 pairs for validation and test, respectively. In total, the dataset has 1,047 categories.

\subsection{Evaluation Protocol}

We evaluate our proposed model with the same metrics used in prior works \cite{salvador2017learning,chen2018deep,carvalho2018cross,zhu2019r2gan,wang2019learning}. To be specific, median retrieval rank (MedR) and recall at top K (R@K) are used. MedR measures the median rank position among where true positives are returned. Therefore, higher performance comes with a lower MedR score. Given a food image, R@K calculates the fraction of times that the correct recipe is found within the top-K retrieved candidates, and vice versa. Different from MedR, the performance is directly proportional to the score of R@K. In the test phase, we first sample 10 different subsets of 1,000 pairs (1k setup), and 10 different subsets of 10,000 (10k setup) pairs. It is the same setting as in \cite{salvador2017learning}. We then consider each item from food image modality in subset as a query, and rank samples from recipe modality according to L2 distance between the embedding of image and that of the recipe, which is served as image-to-recipe retrieval, and vice versa for recipe-to-image retrieval.  

\begin{table*}
  \centering
  \caption{\textbf{Ablation Studies}. Evaluation of benefits of different components of the SCAN framework. The models are evaluated based on MedR, where lower is better, and R@K (\%), where higher is better.}
  {
    \begin{tabular}{clcccc}
    \toprule
    \multicolumn{1}{c}{$L_{Ret}$} & \textbf{Component} & \textbf{medR $\downarrow$} & \textbf{R@1 $\uparrow$} &\textbf{R@5 $\uparrow$}& \textbf{R@10 $\uparrow$} \\
    \midrule
    \multirow{5}{*}{{Cosine Loss}}
    &\textbf{CL} & 2.0   & 46.9  & 76.5  & 84.9 \\
    &\textbf{CL+SA} & 1.0 & 51.0  & 79.9 & 86.3 \\
    \cline{2-6}
    &\textbf{CL+cls} & 1.9 & 47.0  & 75.5  & 83.4 \\
    &\textbf{CL+SC} & 1.0  & 50.7  & 79.7  & 86.4 \\
    \cline{2-6}
    &\textbf{CL+SC+SA} & 1.0  & 52.1  & 80.4  & 87.2 \\
    \midrule
    \multirow{5}{*}{{Triplet Loss}}
    &\textbf{TL} & 2.0   & 47.5  & 76.2  & 85.1 \\
    &\textbf{TL+SA} & 1.0   & 52.5  & 81.1 & 88.4 \\
    \cline{2-6}
    &\textbf{TL+cls} & 1.7 & 48.5  & 78.0  & 85.5 \\
    &\textbf{TL+SC} & 1.0  & 51.9  & 80.3  & 88.0 \\
    \cline{2-6}
    &\textbf{SCAN} & \textbf{1.0} & \textbf{54.0}  & \textbf{81.9}  & \textbf{89.2} \\
    \bottomrule
    \end{tabular}%
  }
  \label{tab:ablation}%
\end{table*}%

\subsection{Implementation Details}
We set the trade-off parameter $\lambda$ in Eq. \eqref{total_eq} based on empirical observations, where we tried a range of values and evaluated the performance on the validation set, as shown in Table \ref{tab:parameter}. We set the $\lambda$ as 0.05. The model was trained using Adam optimizer \cite{kingma2014adam} with the batch size of 64 in all our experiments. The initial learning rate is set as 0.0001, and the learning rate decreases 0.1 in the 30th epoch. We take a pretrained ResNet-50 and the bidirectional LSTMs as $E_I$ and $E_R$ respectively, whose output dimension is $1024$. Note that we update the two sub-networks, i.e. image encoder $E_I$ and recipe encoder $E_R$, alternatively. It only takes 40 epochs to get the best performance with our proposed methods, while \cite{salvador2017learning} requires 220 epochs to converge. Our training records can be viewed in Figure \ref{fig:training}. We do our experiments on a single Tesla V100 GPU, which costs about 16 hours to finish the training.

\subsection{Baselines}
We compare the performance of our proposed methods with several state-of-the-art baselines, and the results are shown in Table ~\ref{tab:results}.

\textbf{CCA \cite{hotelling1936relations}:} Canonical Correlation Analysis (CCA) is one of the most widely-used classic models for learning a common embedding from different feature spaces. CCA learns two linear projections for mapping text and image features to a common space that maximizes their feature correlation. 

\textbf{SAN \cite{chen2017cross}:} Stacked Attention Network (SAN) considers ingredients only (and ignores recipe instructions), and learns the feature space between ingredient and image features via a two-layer deep attention mechanism.

\textbf{JE \cite{salvador2017learning}:} They use pairwise cosine embedding loss to find a joint embedding (JE) between the different modalities. To impose regularization, they add classifiers to the cross-modal embeddings which predict the category of a given food item.

\textbf{AM \cite{chen2018deep}:} Attention mechanism (AM) over the recipe is adopted in \cite{chen2018deep}, applied at different parts of a recipe (title, ingredients and instructions). They use an extra transformation matrix and context vector in the attention model.

\textbf{AdaMine \cite{carvalho2018cross}:} A double triplet loss is used, where triplet loss is applied to both the joint embedding learning and the auxiliary classification task of categorizing the embedding into an appropriate category. They also integrate the adaptive learning schema (AdaMine) into the training phase, which performs adaptive mining for significant triplets.

\textbf{$\mathbf{R^2 GAN}$ \cite{zhu2019r2gan}:} After embedding the image and recipe information, $\rm R^2 GAN$ adopt GAN learning and semantic classification for cross-modal retrieval. They also introduce two-level ranking loss at embedding and image spaces.

\textbf{MCEN \cite{fu2020mcen}:} Fu et al. adopt the generative idea, where they convert the embedding computation into a generative process. They first sample the latent variables from Gaussian distributions, based on which they use several layers to generate new feature embeddings. This method inevitably increase the computational cost.

\textbf{ACME \cite{wang2019learning}:} Adversarial training methods are utilized in ACME for modality alignment, to make the feature distributions from different modalities to be similar. To further preserve the semantic information in the cross-modal food data representation, Wang et al. introduce a translation consistency component.

\textbf{tri-pro \cite{zan2020sentence}:} Zan et al. propose to use the improved triplet loss, where they enforce the embeddings of all images for a given recipe to be close to this recipe, while to be distant from other recipes. They also attempt to discover some hard negatives during training.

\begin{figure*}[h!]
\begin{center}
\includegraphics[width=0.6\textwidth]{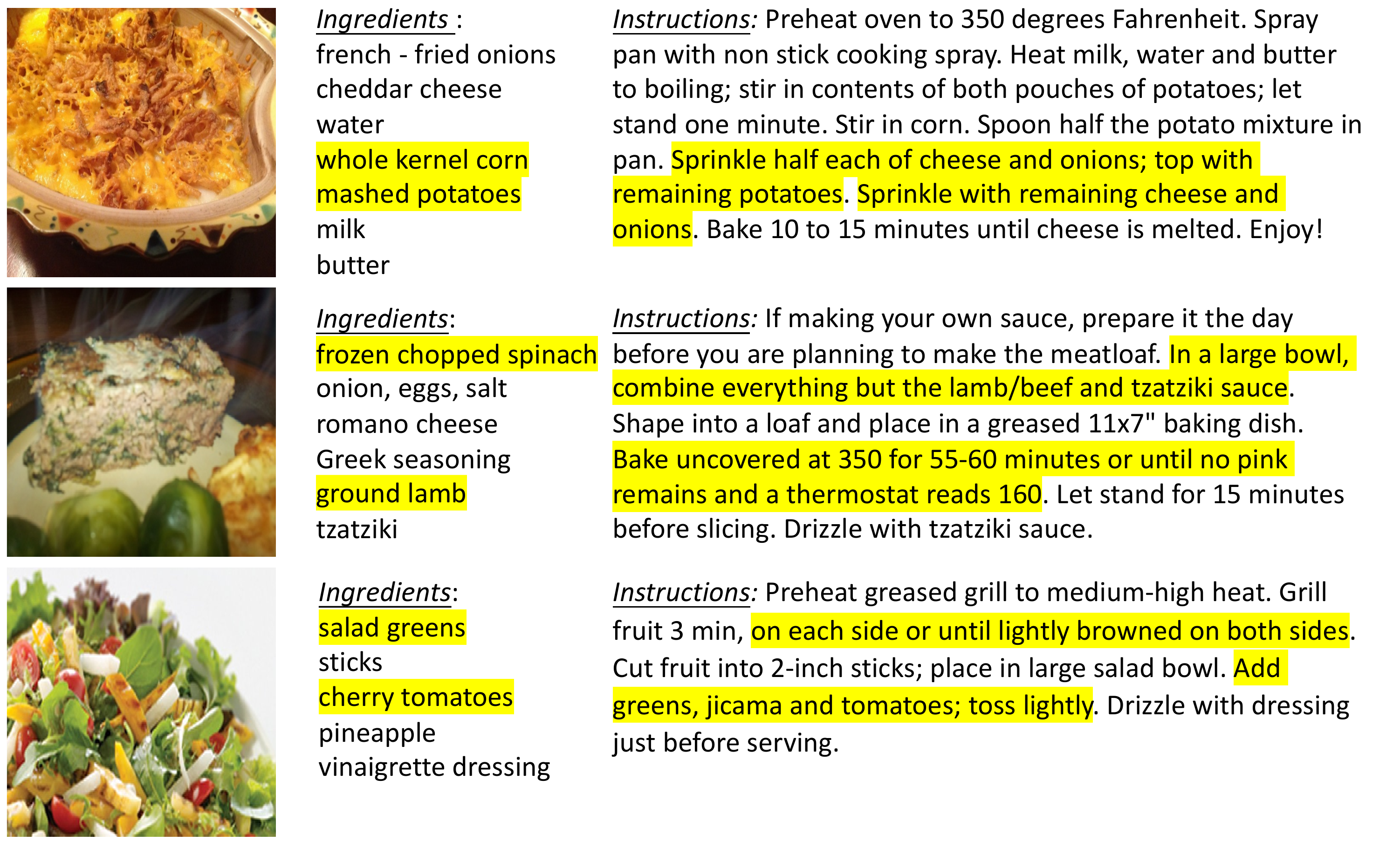}
\end{center}
   \caption{Visualizations of image-to-recipe retrieval. We show the retrieved recipes of the given food images, along with the attended ingredients and cooking instruction sentences.}
\label{fig:attn_vis}
\end{figure*}

It has been validated that using attention can improve feature representations. Both SAN \cite{chen2017cross} and AM \cite{chen2018deep} adopt the attention mechanism to improve the recipe embeddings, while these methods add extra learnable layers to compute the attention weights and need to rely on the high-quality food images, which may affect the model performance. In contrast, without adding extra layers or requiring food images, our adopted self-attention method can find the attended ingredients and cooking instructions effectively. We show some attention results by our model in Figure \ref{fig:attn_vis}. To improve the modality alignment, $\mathbf{R^2 GAN}$ \cite{zhu2019r2gan} and ACME \cite{wang2019learning} use adversarial learning. Specifically, Wang et al. \cite{wang2019learning} preserve the semantic consistency by transforming the feature representations to another modality. While our proposed method can achieve semantic alignment with the KL divergence, which is light-weight and effective. In summary, it can be observed our proposed model SCAN is useful on cross-modal food retrieval and outperforms all of earlier methods, as is shown in Table \ref{tab:results}.

\begin{figure*}
\begin{center}
\includegraphics[width=\textwidth]{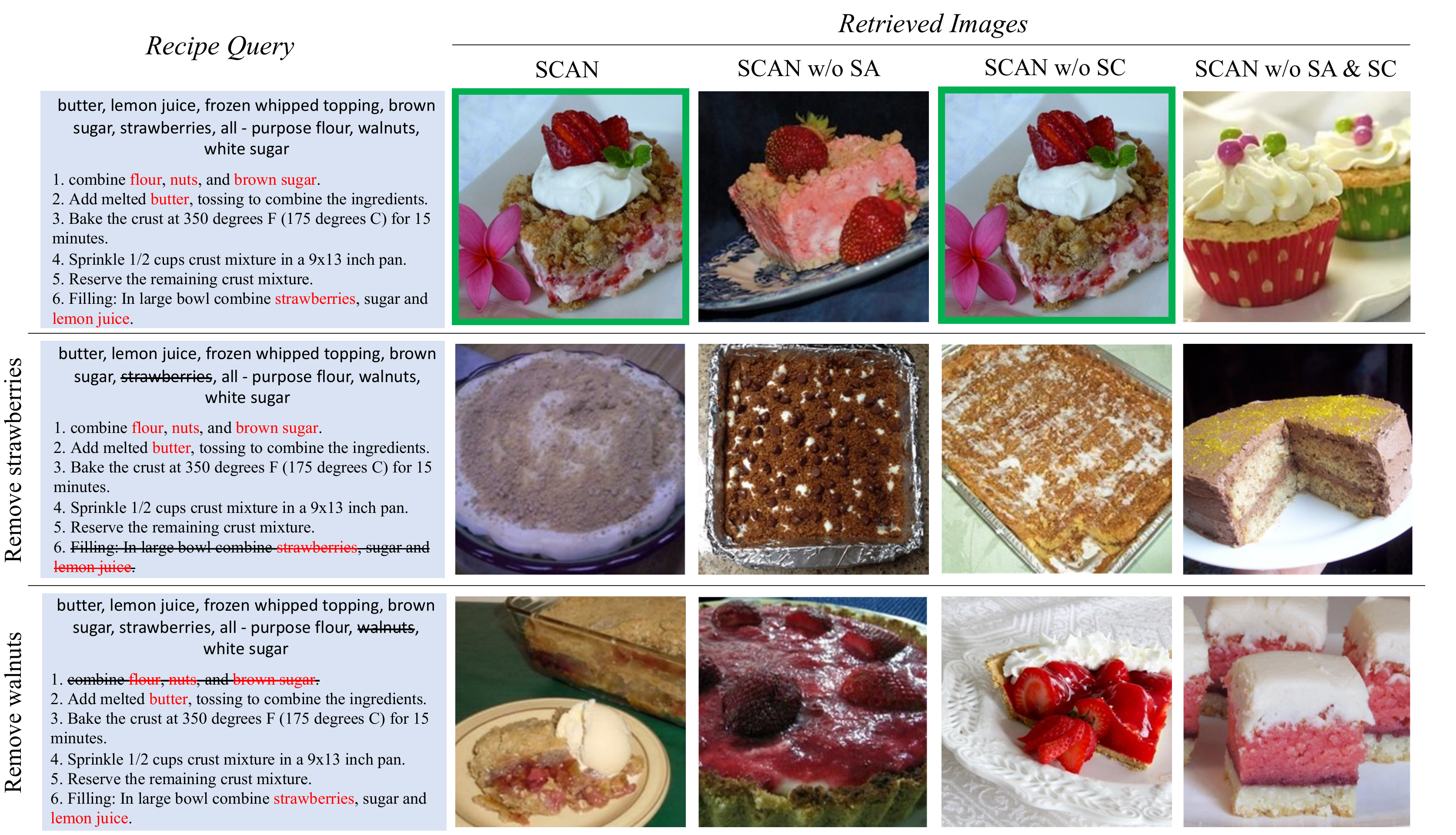}
\end{center}
   \caption{Recipe-to-image retrieval results in Recipe1M dataset. We give an original recipe query from \emph{dessert}, and then remove different ingredients of \emph{strawberries} and \emph{walnuts} separately to construct new recipe queries. We show the retrieved results by SCAN and different components of our proposed model.}
\label{fig:re2im}
\end{figure*}

\subsection{Ablation Studies}
Extensive ablation studies are conducted to evaluate the effectiveness of each component of our proposed model. Table \ref{tab:ablation} illustrates the contributions of self-attention model (\textbf{SA}), semantic consistency loss (\textbf{SC}) and their combination on improving the image to recipe retrieval performance. We test these different components based on different retrieval learning loss functions $L_{Ret}$, i.e. triplet loss (\textbf{TL}) and cosine embedding loss (\textbf{CL}). 

\textbf{TL} serves as a baseline for \textbf{SA}, which adopts the \emph{BatchHard} \cite{hermans2017defense} training strategy. We then add \textbf{SA} and \textbf{SC} incrementally, and significant improvements can be found in both of the two components. To be specific, integrating \textbf{SA} into \textbf{TL} helps improve the performance of the image-to-recipe retrieval more than 4\% in R@1, illustrating the effectiveness of the self-attention mechanism to learn discriminative recipe representations. The model trained with triplet loss and classification loss (\textbf{cls}) used in \cite{salvador2017learning} is another baseline for \textbf{SC}. It shows that our proposed semantic consistency loss improves the performance in R@1 and R@10 by more than 2\%, which suggests that reducing intra-class variance can be helpful in the cross-modal retrieval task. When we add \textbf{SA} and \textbf{SC} to \textbf{CL}, similar improvements can also been observed.

We show the training records in Figure \ref{fig:training}, in the left figure, we can see that for the first 20 epochs, the performance gap between \textbf{TL} and \textbf{TL+SA} gets larger, while the performance of \textbf{TL+cls} and \textbf{TL+SC} keeps being similar, which is shown in the middle figure. But for the last 20 training epochs, the performance of \textbf{TL+SC} improves significantly, which indicates that for those hard samples whose intra-variance can hardly be reduced by \textbf{TL+cls}, \textbf{TL+SC} contributes further to the alignment of paired cross-modal data. The effect of trade-off parameter $\lambda$ is shown in Table \ref{tab:parameter}. We illustrate the performance of models trained with four different $\lambda$, and we can see that setting $\lambda$ as 0.05 can obtain the best performance.

In conclusion, we observe that each of the proposed components improves the cross-modal retrieval model, and the combination of those components yields better performance overall. 

\subsection{Recipe-to-Image Retrieval Results}
We show three recipe-to-image retrieval results in Figure \ref{fig:re2im}. In the top row, we select a recipe query \emph{dessert} from Recipe1M dataset, which has the ground truth for retrieved food images. Images with the green box are the correctly retrieved ones, which come from the retrieved results by SCAN and TL+SA. But we can see that the model trained only with semantic consistency loss (TL+SC) has a reasonable retrieved result as well, which is relevant to the recipe query.

In the middle and bottom row, we remove some ingredients and the corresponding cooking instruction sentences in the recipe, and then construct the new recipe embeddings for the recipe-to-image retrieval. In the bottom row where we remove the \emph{walnuts}, we can see that all of the retrieved images have no \emph{walnuts}. However, only the image retrieved by our proposed SCAN reflects the richest recipe information. For instance, the image from SCAN remains visible ingredients of \emph{frozen whipped topping}, while images from TL+SC and TL+SA have no \emph{frozen whipped toppings}. 

The recipe-to-image retrieval results indicate an interesting way to satisfy users' needs to find the corresponding food images for their customized recipes.

\subsection{Image-to-Recipe Retrieval Results \& Effect of Self-Attention model}\label{attention_vis}
In this section, we show some of the image-to-recipe retrieval results in Figure \ref{fig:attn_vis} and then focus on analyzing the effect of our self-attention model. Given images from \emph{cheese cake}, \emph{meat loaf} and \emph{salad}, we show the retrieved recipe results by SCAN, which are all correct. We visualize the attended ingredients and instructions for the retrieved recipes with the yellow background, where we choose the ingredients and cooking instruction sentences of the top 2 attention weights as the attended ones. We can see that some frequently used ingredients like \emph{water}, \emph{milk}, \emph{salt}, etc. are not attended with high weights, since they are not visible and shared by many kinds of food, which cannot provide enough discriminative information for cross-modal food retrieval. This is an intuitive explanation for the effectiveness of our self-attention model. 

Another advantage of using the self-attention mechanism is that the image quality cannot affect the attended outputs. Obviously, the top two rows of food images \emph{cheese cake} and \emph{meat loaf} do not have good image quality, while our self-attention model still outputs reasonable attended results. This suggests that our proposed attention model has good capabilities to capture informative and reasonable parts for recipe embedding. 

\begin{figure}[h!]
\begin{center}
\includegraphics[width=0.5\textwidth]{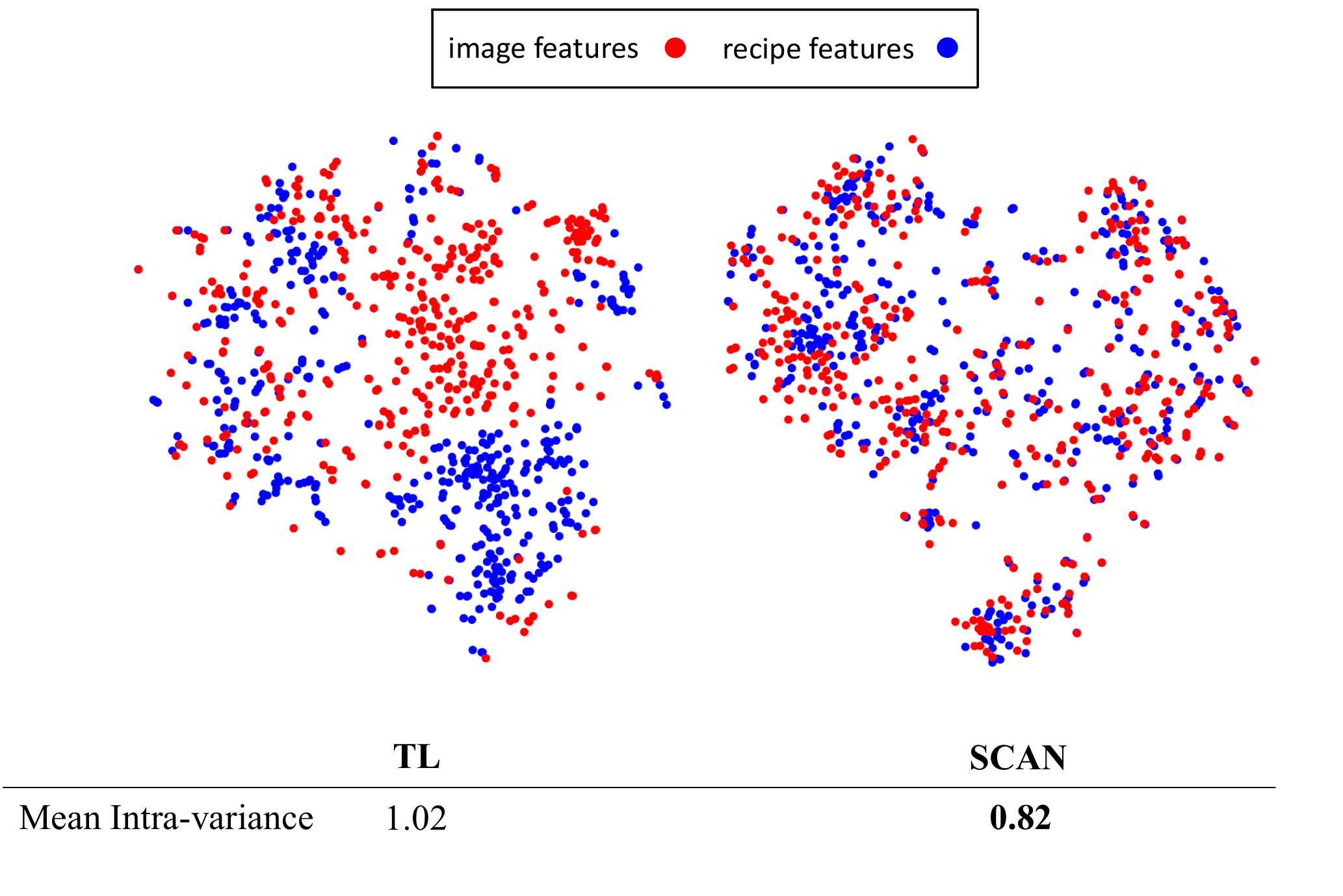}
\end{center}
   \caption{The difference on the intra-class feature distance of cross-modal paired data trained without and with semantic consistency loss. The food data is selected from the same category, \emph{chocolate chip}. \textbf{SCAN} obtains closer image-recipe feature distance than \textbf{TL}. (Best viewed in color.)}
\label{fig:SCC_vis}
\end{figure}

\subsection{Effect of Semantic Consistency}

To have a concrete understanding of the ability of our proposed semantic consistency loss on reducing the mean intra-class feature distance (intra-class variance) between paired food image and recipe representations, we show the difference of the intra-class feature distance on cross-modal data trained without and with semantic consistency loss, i.e. TL and SCAN, in Figure \ref{fig:SCC_vis}. In the test set, we select the recipe and food image data from \emph{chocolate chip}, which in total has 425 pairs. We obtain the food data representations from models trained with two different methods, then we compute the Euclidean distance between paired cross-modal data to obtain the mean intra-class feature distance. We adopt t-SNE \cite{maaten2008visualizing} to do dimensionality reduction to visualize the food data.

It can be observed that cross-modal food data which is trained with semantic consistency loss (SCAN) has smaller intra-class variance than that trained without semantic consistency loss (TL). This means that semantic consistency loss is able to correlate paired cross-modal data representations effectively by reducing the intra-class feature distance, and also our experiment results suggest its efficacy.

\section{Conclusion}

In conclusion, we propose SCAN, an effective training framework, for cross-modal food retrieval. It introduces a novel semantic consistency loss and employs the self-attention mechanism to learn the joint embedding between food images and recipes for the first time. To be specific, we apply semantic consistency loss to cross-modal food data pairs to reduce the intra-class variance, and utilize the self-attention mechanism to find the important parts in the recipes to construct discriminative recipe representations. SCAN is easy to implement and can extend to other general cross-modal datasets. We have conducted extensive experiments and ablation studies. We achieved state-of-the-art results in Recipe1M dataset.

\section*{Acknowledgement}
This research is supported by the National Research Foundation, Singapore under its International Research Centres in Singapore Funding Initiative. Any opinions, findings and conclusions or recommendations expressed in this material are those of the author(s) and do not reflect the views of National Research Foundation, Singapore.

% \newpage

\bibliographystyle{IEEEtran}
\bibliography{IEEEexample}



% \end{thebibliography}

% \begin{IEEEbiography}[{\includegraphics[width=1in,height=1.25in,clip,keepaspectratio]{mshell}}]{Michael Shell}

\newpage

\begin{IEEEbiography}[{\includegraphics[width=1in,height=1.25in,clip,keepaspectratio]{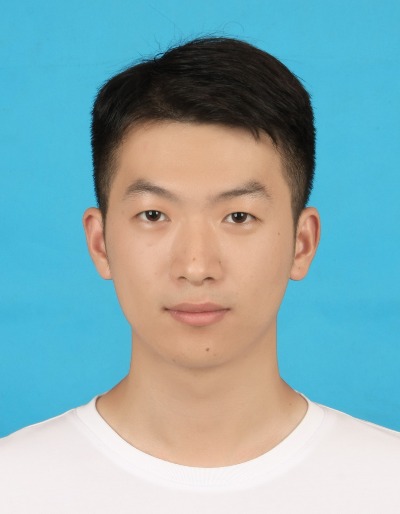}}]{Hao Wang}
is a PhD candidate with the School of Computer Science and Engineering, Nanyang Technological University, Singapore. His research interests include multi-modal analysis and computer vision.
\end{IEEEbiography}

\begin{IEEEbiography}[{\includegraphics[width=1in,height=1.25in,clip,keepaspectratio]{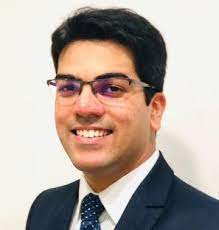}}]{Doyen Sahoo}
is a Senior Research Scientist at Salesforce Research Asia. Prior to joining Salesforce, Doyen was a Research Fellow at the Living Analytics Research Center at Singapore Management University (SMU). He was also serving as Adjunct Faculty in SMU. Doyen earned his PhD in Information Systems from SMU in 2018 and B.Eng in Computer Science from Nanyang Technological University in 2012. His research interests include Online Learning, Deep Learning, Computer Vision, and he also works on applied research including AIOps, Computational Finance and Cyber Security applications. He has published over 40 articles in top tier conferences and journals including ICLR, CVPR, ACL, KDD, JMLR, etc.
\end{IEEEbiography}

\begin{IEEEbiography}[{\includegraphics[width=1in,height=1.25in,clip,keepaspectratio]{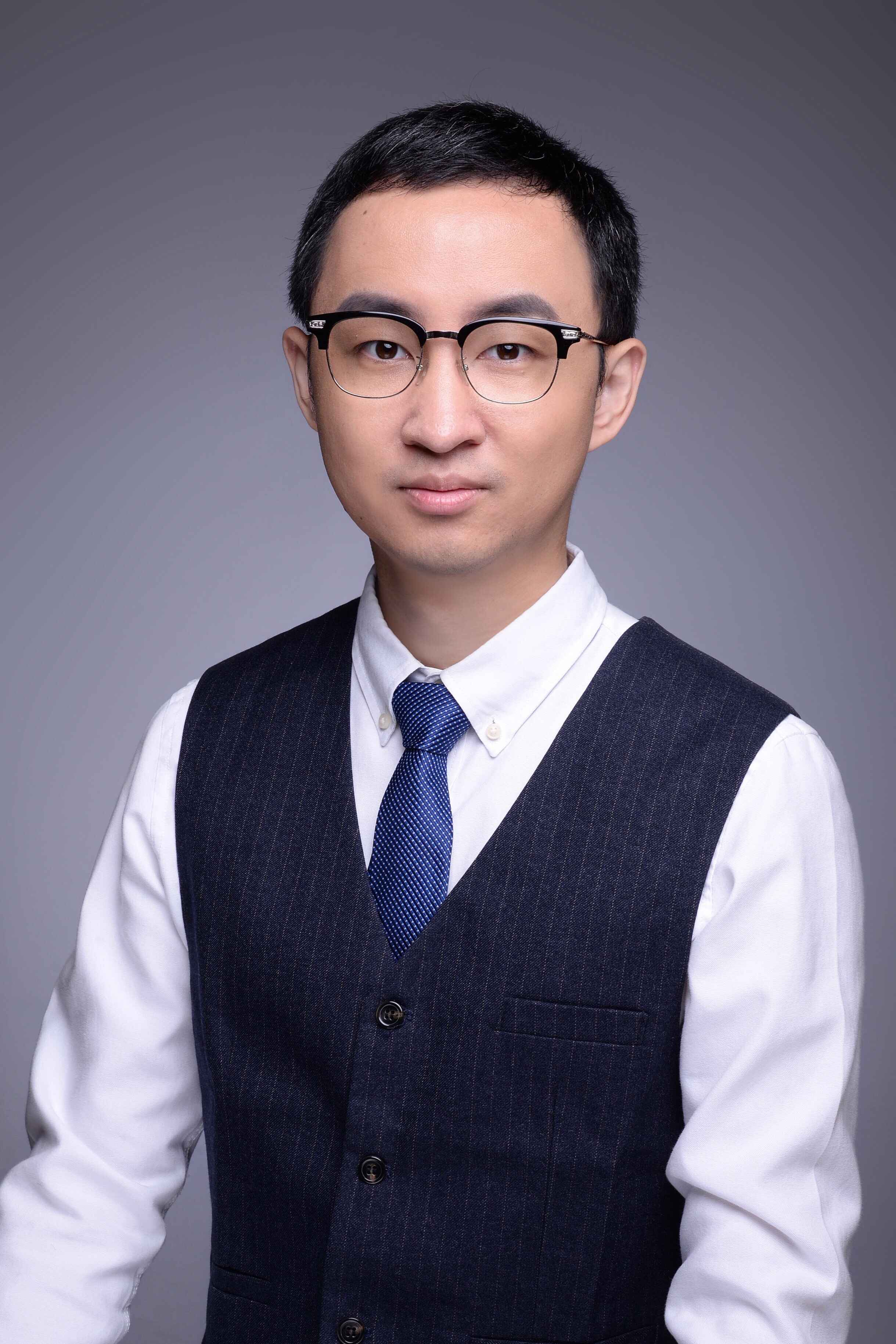}}]{Chenghao Liu}
is currently a senior applied scientist of Salesforce Research Asia. Before, he was a research scientist in the School of Information Systems (SIS), Singapore Management University (SMU), Singapore. He received his Bachelor degree and Ph.D degrees from the Zhejiang University. His research interests include large-scale machine learning (online learning and deep learning) with application to tackle big data analytics challenges across a wide range of real-world applications.
\end{IEEEbiography}

\begin{IEEEbiography}[{\includegraphics[width=1in,height=1.25in,clip,keepaspectratio]{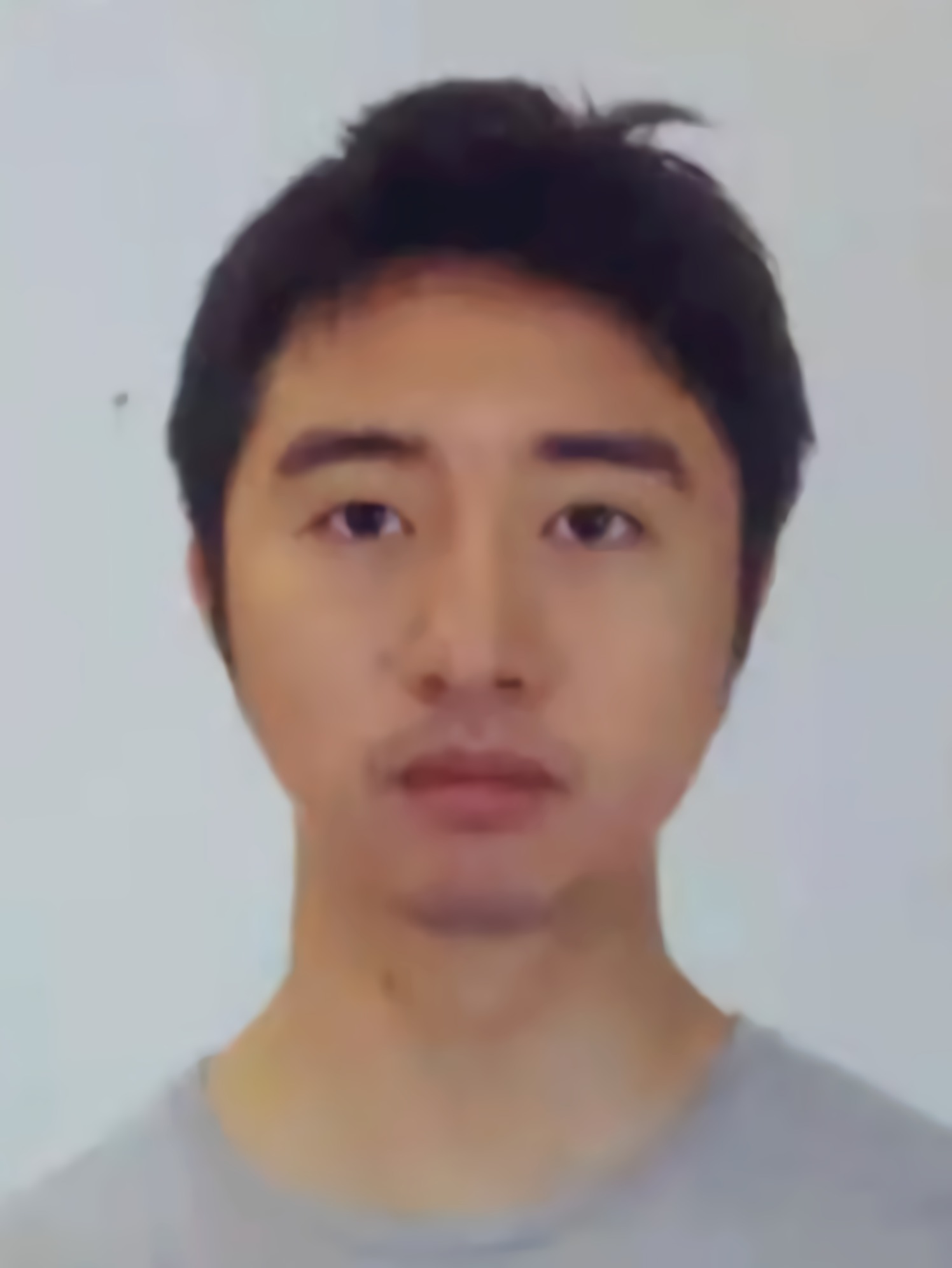}}]{Ke Shu}
is a research engineer at the Living Analytics Research Centre (LARC), Singapore Management University. His research interests include machine learning and deep learning.
\end{IEEEbiography}

\begin{IEEEbiography}[{\includegraphics[width=1in,height=1.25in,clip,keepaspectratio]{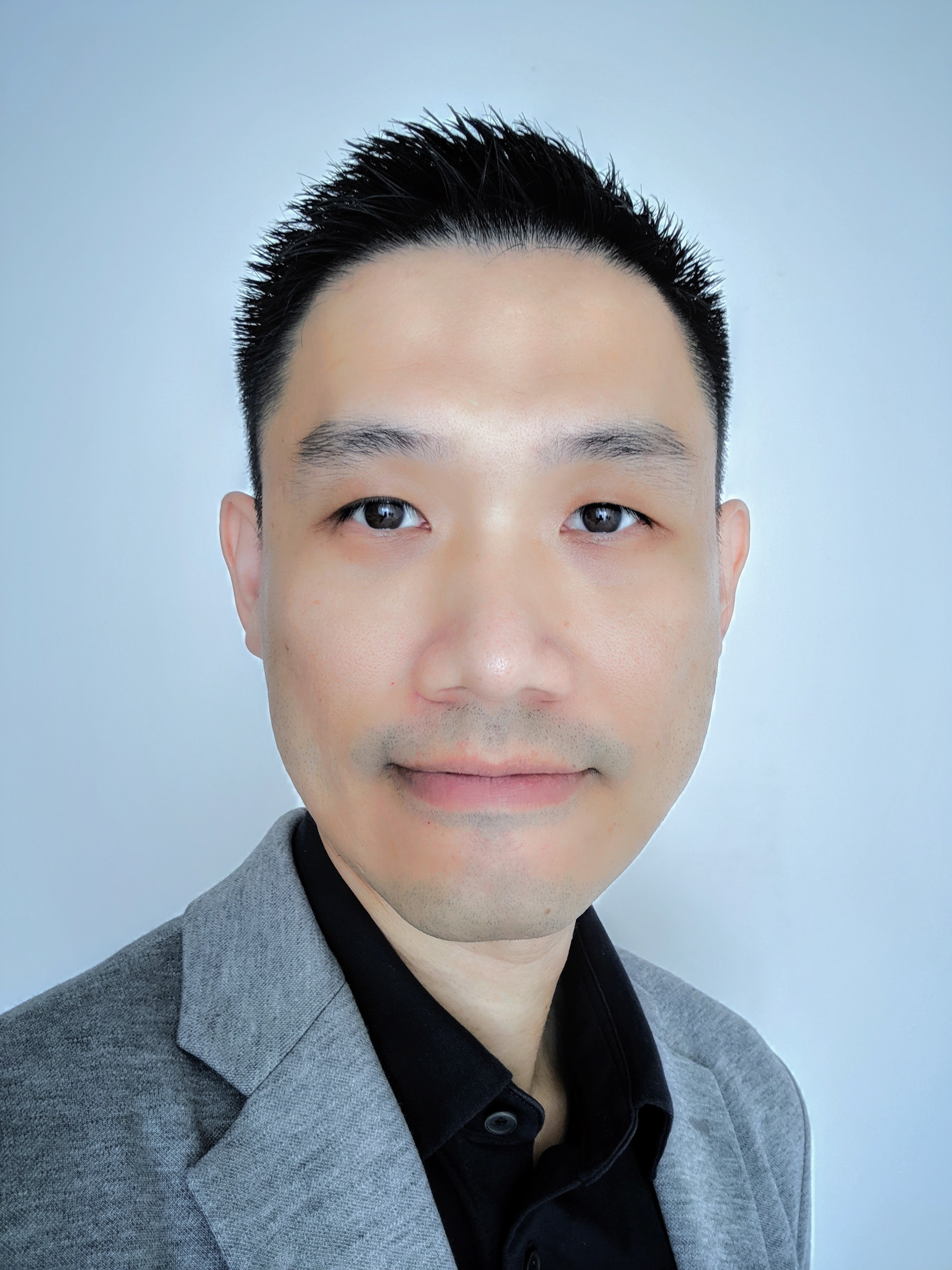}}]{Palakorn Achananuparp}
is a senior research scientist at the Living Analytics Research Centre (LARC), Singapore Management University. He is interested in developing and applying machine learning, natural language processing, and crowdsourcing techniques to solve problems in a variety of domains, including online social networks, politics, and public health.
\end{IEEEbiography}

\begin{IEEEbiography}[{\includegraphics[width=1in,height=1.25in,clip,keepaspectratio]{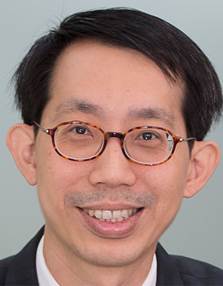}}]{Ee-peng Lim}
is the Lee Kong Chian Professor with the School of Computing and Information Systems at the Singapore Management University. He is also the Director of Living Analytics Research Centre in the School, a research centre focusing developing personalized and participatory analytics capabilities for smart city and smart nation relevant applications. Dr Lim received his PhD degree from University of Minnesota. His research expertise covers social media mining, social/urban data analytics, and information retrieval. He is the recipient of the Distinguished Contribution Award at the 2019 Pacific Asia Conference on Knowledge Discovery and Data Mining (PAKDD), and the Test of Time award at 2020 ACM Conference on Web Search and Data Mining (WSDM).
\end{IEEEbiography}

\begin{IEEEbiography}[{\includegraphics[width=1in,height=1.25in,clip,keepaspectratio]{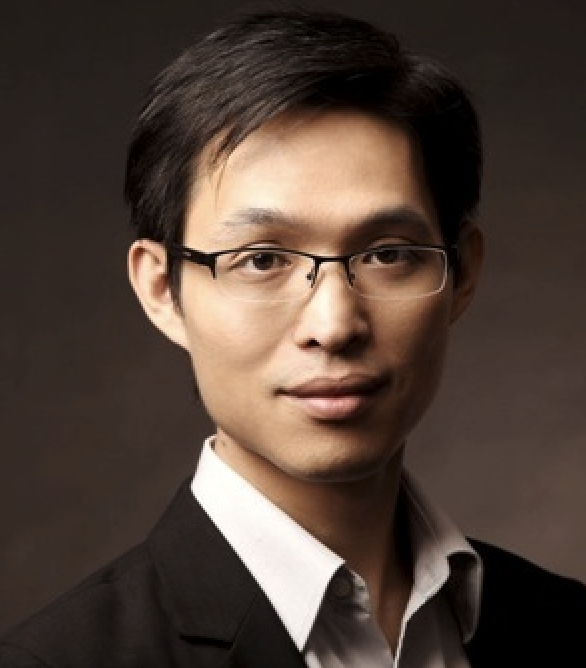}}]{Steven C. H. Hoi}
is currently the Managing Director of Salesforce Research Asia, and a Professor of Information Systems at Singapore Management University, Singapore. Prior to joining SMU, he was an Associate Professor with Nanyang Technological University, Singapore. He received his Bachelor degree from Tsinghua University, P.R. China, in 2002, and his Ph.D degree in computer science and engineering from The Chinese University of Hong Kong, in 2006. His research interests are machine learning and data mining and their applications to multimedia information retrieval, social media and web mining, and computational finance, etc. He has served as the Editor-in-Chief for Neurocomputing Journal, general co-chair for ACM SIGMM Workshops on Social Media, program co-chair for the fourth Asian Conference on Machine Learning, book editor for “Social Media Modeling and Computing”, guest editor for ACM Transactions on Intelligent Systems and Technology. He is an IEEE Fellow and ACM Distinguished Member.
\end{IEEEbiography}


% % insert where needed to balance the two columns on the last page with
% %\newpage



%\vfill


\end{document}